\documentclass[sigconf]{acmart}

\setcopyright{acmlicensed}
\copyrightyear{2025}
\acmYear{2025}

\acmDOI{10.48550/arXiv.2511.10659}

\acmWorkshop[ACM ICAIF '25]{Being Presented at the AI for Financial Inclusion, Risk Modeling and Resilience in Emerging Markets
workshop at ACM ICAIF 2025 Singapore}{November 15--18, 2025}{Singapore}

\usepackage{multirow}

\begin{document}

\title{Information Extraction From Fiscal Documents Using LLMs}


\author{Vikram Aggarwal}
\email{viki@google.com}
\orcid{0009-0004-1750-6806}
\affiliation{%
  \institution{Google Inc}
  \city{Mountain View}
  \country{USA}
}

\author{Jay Kulkarni}
\authornotemark[1]
\email{jayvilaskulkarni@gmail.com}
\affiliation{%
	\institution{XKDR Forum}
	\city{Mumbai}
	\country{India}
}

\author{Aditi Mascarenhas}
\authornotemark[1]
\email{mascarenhasaditi@gmail.com}
\affiliation{%
	\institution{XKDR Forum}
	\city{Mumbai}
	\country{India}
}

\author{Aakriti Narang}
\authornotemark[1]
\email{narangaakriti9@gmail.com}
\affiliation{%
	\institution{XKDR Forum}
	\city{Mumbai}
	\country{India}
}

\author{Siddarth Raman}
\authornotemark[1]
\email{raman.siddarth@gmail.com}
\affiliation{%
  \institution{XKDR Forum}
  \city{Mumbai}
  \country{India}
}

\author{Ajay Shah}
\authornotemark[1]
\email{ajayshah@mayin.org}
\affiliation{%
  \institution{XKDR Forum}
  \city{Mumbai}
  \country{India}
}

\author{Susan Thomas}
\authornotemark[1]
\email{sthomas.entp@gmail.com}
\affiliation{%
  \institution{XKDR Forum}
  \city{Mumbai}
  \country{India}
}


\renewcommand{\shortauthors}{Aggarwal et al.}


\begin{abstract}
  Large Language Models (LLMs) have demonstrated remarkable
  capabilities in text comprehension, but their ability to process
  complex, hierarchical tabular data remains underdeveloped.  We
  present a novel approach to extracting structured data from
  multi-page government fiscal documents using LLM-based techniques.
  Applied to large fiscal documents from the State of Karnataka
  in India, our method achieves high accuracy by combining domain knowledge, sequential context, and algorithmic validation. Traditional OCR methods work poorly to translate information from PDF documents into a structured database of fiscal information, with errors that are hard to detect. The inherent structure within fiscal documents, with totals at each level of the hierarchy, opens up the possibility to extract information, with 
  robust internal validation of the extracted data. 
  We demonstrate that LLMs can read tables and also process
  document-specific structural hierarchies, offering a 
  process for converting PDF-based fiscal disclosures into
  research-ready databases that can be scaled up. We use the
  hierarchical relationships within fiscal documents to create multi-level validation checks. Our implementation shows promise for
  broader applications involving complex tables, particularly for developing country contexts.
\end{abstract}



\maketitle

\section{Introduction}

While LLMs excel at processing natural language, their ability to
process tabular data has not been as emphasised or understood. In the real world, a great deal of financial disclosures, bills, and fiscal documents are
published in PDF formats containing tables of data that have inherent, complex relationships. These are non-linear in the flow of information, that are not easily machine-analysable. While the source information exists in the PDF,
data analysis requires human knowledge to understand the document
structure and parse it correctly. A simple extraction of
information from PDF files is particularly prone to errors when tables have inconsistent presentation of text and numbers. Research in this field is 
hampered by the lack of a large corpus of PDF documents and their
associated parsed, structured text.

In India, complex hierarchical public finance databases are 
released as large PDF files. While the PDF files contain all the
information and are readily available, many applications require the
underlying database, and this is hard to extract. In this paper, we
show an LLM-based process that is able to map the public domain PDF
files into accurate datasets. While the ideas are general, our
implementation is done using the fiscal documents of the Indian state of Karnataka. For this setting:
\begin{itemize}
\item We show that the format of information in the fiscal documentation of Karnataka is representative of the complexity of fiscal disclosures of Indian states. 
\item We develop a process for extracting machine-analysable
  information from the fiscal documents of this state.
\item We demonstrate how to identify the accuracy of the information
  extracted, even when no ground truth dataset exists.  Relying on the
  structure of fiscal documents, our mechanism also identifies the
  locations where the information extraction failed.
\end{itemize}


\subsection{The Challenge}

Sophisticated applications of public finance data require analysis
applied to the underlying dataset. Economists, civil society
organizations and journalists can use this data to tag different heads
of expenditure, and categorize spending by function, geography,
department, and other dimensions.  This allows for an analysis of
government spending patterns, efficiency, and shifts in priorities
over time. Examples of applications include:
\begin{itemize}
  \item Public procurement analysis
  \item Budgeting efficiency studies (Budgeted vs Revised vs Actuals)
  \item Expenditure priority tracking over time
  \item In-year fund reallocation analysis
  \item Functional/departmental/geographic classification
\end{itemize}

When the public finance dataset is represented as a PDF file, this
hinders analysis. Traditional OCR approaches are not effective in
recovering the underlying data because:

\begin{itemize}
  \item Tables span multiple pages; tables have varying structures.
  \item The files are large: roughly hundreds of pages.  Since
    tables span multiple pages, page-by-page extraction missed vital
    information.
  \item In India, the documents feature an inconsistent mix of English
    and regional languages.
  \item Units and formatting vary across documents.
\end{itemize}

For researchers and civil society organisations aiming to analyse
public finance, this creates a significant barrier. Manual data entry
is labor-intensive and error-prone. Converting these documents into
structured, machine-readable formats through a mix of text and image
processing tools is a laborious task that requires manual intervention
and verification, as well as domain expertise with public finance. The need for such effort acts as a barrier to scaling up both the availability of structured databases of fiscal information, as well as the analysis that is possible when such large scale databases are available.


\subsection{The Opportunity}

Unlike general text extraction, the deep hierarchical fiscal documents
contain a natural validation mechanism: columns must sum to reported
totals at upper hierarchical levels. This creates a unique
opportunity to develop verifiable LLM-based extraction systems where
extraction accuracy can be algorithmically validated.

It is a big step forward for data extraction when structured data is
extracted with accuracy scores and identification of the precise
locations of errors. Downstream users could potentially expend a small burden of additional effort to solve the last mile of cleaning, and then arrive at a perfect dataset.

This work may have important implications for future refinement of
LLMs themselves. LLM performance on multi-lingual, hierarchical,
multi-page tabular data is known to be poor. Our research program can
yield a real world corpus of tens of thousands of pages of parallel
PDF and structured-text data, which can be used for training
LLMs. When such a dataset is created, it will induce improvements in
LLMs for processing such tabular information.

\section{LLM approaches}

LLMs have been used for data extraction and data summarisation, and
for Q\&A tasks.  Early work by Dong et al.~\cite{pretraining22}
covered common formats like web documents, spreadsheets and CSVs, but
PDFs were either under represented, or entirely missing. A recent
survey on small, English language tables from Wikipedia by Sui et
al.~\cite{tablemeetsllm} finds that LLMs do not have a good
understanding of table structure.  Currently, most research LLMs
focuses on understanding tabular data primarily for the question
answering task. Most papers (ex. Zhao et
al.\cite{zhao-etal-2023-large}, Bhandari et
al.~\cite{llmrobustness24}) focus on machine readable formats like
JSON, or text formats delimited by pipe characters.  Even working with
simpler formats, Zhao et al.~\cite{zhao-etal-2023-large} demonstrate
that tabular data with hierarchical information is poorly understood
by current LLMs. Lu et al.~\cite{Lu2025} published a survey of LLM
techniques showing that academic research has been focused on easier,
text-based document types (text, JSON, HTML), and has not fully
investigated image, or page-based formats like PDF, and has not fully
explored real-world complexity: multi-page tables, multi-lingual
tables, hierarchical tables, and PDF files spanning multiple pages.
The authors conclude, ``\textit{Future table LLMs should adapt quickly and
cheaply to real world business needs.  Research directions include
synthesizing high-quality training data that reflects the diverse
needs of specific domains by cost-effective methods.}''

There have been efforts to create evaluation datasets
(ex.~\cite{tableeval25, rdtablebench, tablebench25}) with complicated
tables representative of real-world use-cases.  At its most complex,
the TableEval~\cite{tableeval25} dataset has dual lingual (English and
Chinese), hierarchical tables generated from computer spreadsheets
provided as JSON or HTML tables.  The largest table in that dataset
(table 520) has 113 rows and 3 columns, and is a 10 kilobyte
spreadsheet. TableEval is intended for Q\&A tasks rather than direct
information extraction.

In contrast, the real world problems that we face are much more
complex. The tables in Indian public finance data have evolved over multiple decades, and are published in the public domain as
PDF files, each containing hundreds of pages.  There are many
different tables of varying layouts, and tables can spread across
pages and contain hundreds of rows. They have interspersed English and
Indic characters with inconsistent coding.

\section{Methodology}

\subsection{Key Innovations}

While straightforward prompting can work for simple tables, complex
hierarchical tables like fiscal documents require additional
techniques to ensure accuracy and consistency.  To ensure high-quality
extraction in a format that is also algorithmically verifiable, we
introduce the following innovations:
\begin{enumerate}
\item \textbf{Image-based processing}: Converting PDF pages to
  high-resolution JPGs (300 DPI) improves LLM comprehension compared
  to direct PDF input or text-based OCR. Converting a document to
  image removes all text metadata, which then forces the LLM to
  perform OCR to recognize text terms, rather than reading the
  (potentially erroneous) text metadata. The LLM may also utilise
  information represented in the physical placement of rows and
  columns in a table. At the same time, working with image files for
  each page introduces new complexities where the context from
  previous pages needs to be carried forward manually.

\item \textbf{Sequential context}: The PDF files that we face have
  hundreds of pages, and rapidly run into the limited context
  window. Hence, each page receives the previous page's extracted data
  as context, enabling state carry-forward across page boundaries.

\item \textbf{Multi-level validation}: Algorithmic checks verify
  summation consistency at the various levels of fiscal reporting:
  Object Head, Detailed Head, Sub Head, Minor Head, Sub Major Head and
  Major Head levels.

\item \textbf{Meta-prompting}: Our initial prompt provides domain
  context about fiscal documents, and provide few-shot examples to the
  LLM, to get the LLM to write the extraction prompt.

\item \textbf{Intelligent cleaning}: A semantic CSV cleaner uses
  row-type understanding (Header/Data/Total) to detect and correct
  column misalignment.
\end{enumerate}

\section{Implementation: Karnataka Finances 2020-21}

\subsection{Precise Problem Statement}

The input PDF files have the following features:
\begin{enumerate}
\item \textbf {Multi-lingual source}: India is a vast, diverse country
  with 22 official languages written in different scripts. States
  conduct business in English, in addition to an official state
  language.  Fiscal information is usually published in English, but
  can also carry the state's official language.  In some cases, the
  source documents are only published in the state's language, and not
  in English.  Further, Indian languages are ``low-resource''
  languages, ie, they suffer from lack of a high quality corpus on the
  web. This leads to worse LLM performance on both information
  extraction and generation tasks, e.g. as demonstrated by Singh
  et. al~\cite{indicgenbench24}.
\item \textbf {Document Encoding}: Since the documents are published
  as PDFs, they can carry font information. We find that the regional
  language is often encoded incorrectly, or is encoded as ASCII and
  font code books are used to display the regional language.
  Information extraction from such malformed documents is difficult,
  as machine extraction ``sees'' ASCII in the metadata.  This is true
  of both conventional metadata-extraction tools (pdf2text) and
  language models. Hence, working with image files is essential, which
  has its own consequences.
\item \textbf {Document Size}: PDF files with more than 500 pages are common. As
  an example, Table~\ref{tab:tedvalidation} on
  page~\pageref{tab:tedvalidation} shows the number of pages in the
  PDF of files that we have analyzed. An information extraction
  technique has to handle this immense size.  As mentioned above, a
  na\"ive approach involving LLMs chokes on overflowing the context
  window.
\item \textbf {Table Structure}: The source documents contain tables
  of different types. The initial tables contain the top-level
  hierarchy, and overall totals. Subsequent tables expand on this with
  department-level detail. LLM prompts need to explain this structure.
\item \textbf {Table Locations}: Fiscal tables often span many pages.
  Page headers and footers have to be ignored, and the extraction
  process needs to remember where a table starts and ends.  Further,
  pages can contain multiple tables, as a single page can end a large
  table from prior pages, fully contain one or more small tables, and
  start another table that continues for many pages.  Any single-page
  extraction prompt needs to consider this level of complexity. While
  most pages are in portrait orientation, sometimes wide tables are
  rotated, and so intervening pages are in landscape orientation.
\item \textbf{Non-standard Number Formatting}: The files use lakhs
  ($10^5$), crores ($10^7$), thousands, mixed use of commas and
  periods, and sometimes non-Arabic numerals.
\item \textbf{Scanned Documents}: Some documents are scanned images,
  others are digitally generated PDFs.
\item \textbf{Hierarchical Tables}: There are multiple nested
  hierarchies, and not all levels are consistent within or across
  states.
\item \textbf{Inconsistent Formatting}: Merged cells, inconsistent
  spacing, additional textual context in the middle of tables that may
  require semantic understanding to parse correctly.
\item \textbf{Repeated Values}: Similar numbers repeated in different
  categories (Revenue / Capital, Voted / Charged).
\end{enumerate}

Each Indian state releases public finance data through PDF files which have some subset of these 10 elements of challenge. The state of Karnataka is a particularly difficult problem, featuring 8 out of 10 of these challenges (i.e. everything enumerated above other than problems 6 and 7).

\subsection{Approach}

\begin{figure}[htbp]
  \centering
  \includegraphics[width=0.3\textwidth]{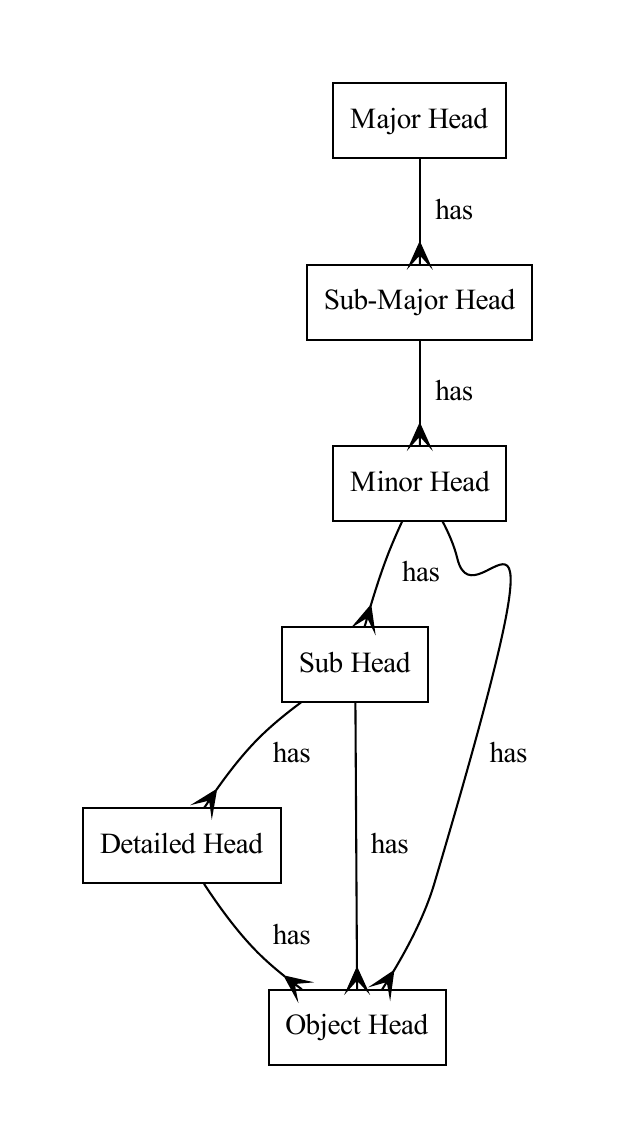}
  \caption{Hierarchy in Karnataka finances}
  \label{fig:hierarchy}
\end{figure}

Our method works in the following steps:
\begin{enumerate}
\item \textbf{Document structure} We provide the LLM with 
  knowledge about the accounting structures specific to Indian government fiscal documents, including
  definitions of Major Head, Minor Head, Object Head, and the typical
  hierarchical relationships. This hierarchy relationship is
  visualised in Figure~\ref{fig:hierarchy} on
  page~\pageref{fig:hierarchy}. The LLM is also provided the full PDF
  document to get the various table types and nesting hierarchies.
\item \textbf{Table structure} We take the document structure, and
  prompt the LLM to create CSV schema for each type of table,
  specifying column names, data types, and expected formats for
  numerical and categorical data.
\item \textbf{Meta-prompt strategy} We give the LLM illustrative pages
  from the document, with the CSV schema from Step 2, and ask it to
  generate the actual extraction prompt.
\end{enumerate}

This prompt is then used to extract each page of the document
sequentially, passing the extracted data from the previous page as
context for the current page.  For each page, the LLM identifies the
table type, applies the relevant schema, and extracts the data into
that specific CSV format.  If there are multiple table types on a
single page, the LLM segments the page accordingly and applies the
appropriate schema to each segment.

\subsection{Output Schema}

For Karnataka, the LLM generates five CSV archetypes:
\begin{enumerate}
  \item Sub-Major Head
  \item Minor Head
  \item Sub Head
  \item Detailed Head
  \item Object Head
\end{enumerate}

\subsection{Cleaning}

The raw LLM output contains misalignment due to merged cells,
inconsistent spacing, and occasional misclassifications of row
types. We implement a semantic CSV cleaner that uses the understanding
of row types (Header/Data/Total) and row levels to detect and correct
column misalignment. This cleaner ensures that each row adheres to
the expected schema, correcting for common errors such as shifted
columns or misplaced totals.

\subsection{Validation Results}

Since we do not have the ground truth, we perform two types of internal
consistency checks as a validation of the extraction process:
\begin{itemize}
\item \textbf{Numerical Consistency}: By verifying sums of Object
  Heads against corresponding Detailed Head totals, we ensure
  within-schema consistency. By matching Minor Head totals from two
  different schema we ensure across-schema consistency. It is
  possible to perform the within- and across- schema numerical
  consistency checks at all levels of hierarchy.
\item \textbf{Structural Soundness}: We calculate the Tree Edit
  Distance Similarity (TEDS), an established metric from Zhong et
  al.~\cite{pubtabnet2020} to compare hierarchical structure. TEDS
  measures how accurately fiscal heads across different schema match
  identical structures at the same depth.  Here, the numbers are
  irrelevant, and we're looking to match the hierarchical structure
  from various sources to validate the extraction process.
\end{itemize}

\begin{table}[h]
\centering
\small
\begin{tabular}{clrrr}
\toprule
\textbf{Vol.}    & \textbf{Validation Type} & \textbf{Checks} & \textbf{Passed} & \textbf{Pass Rate \%} \\
\midrule
\multirow{3}{*}{1} & Object → Detailed Head   & 374             & 317             & 85\%                  \\
                   & Two-source Minor Head    & 154             & 146             & 95\%                  \\
                   & Overall                  & 528             & 463             & 88\%                  \\
\midrule
\multirow{3}{*}{2} & Object → Detailed Head   & 337             & 284             & 84\%                  \\
                   & Two-source Minor Head    & 126             & 118             & 94\%                  \\
                   & Overall                  & 463             & 402             & 87\%                  \\
\midrule
\multirow{3}{*}{3} & Object → Detailed Head   & 199             & 149             & 75\%                  \\
                   & Two-source Minor Head    & 90              & 84              & 93\%                  \\
                   & Overall                  & 289             & 233             & 81\%                  \\
\midrule
\multirow{3}{*}{4} & Object → Detailed Head   & 189             & 147             & 78\%                  \\
                   & Two-source Minor Head    & 60              & 59              & 98\%                  \\
                   & Overall                  & 249             & 206             & 83\%                  \\
\midrule
\multirow{3}{*}{5} & Object → Detailed Head   & 275             & 203             & 74\%                  \\
                   & Two-source Minor Head    & 115             & 113             & 98\%                  \\
                   & Overall                  & 390             & 316             & 81\%                  \\
\midrule
\multirow{3}{*}{6} & Object → Detailed Head   & 99              & 79              & 80\%                  \\
                   & Two-source Minor Head    & 56              & 55              & 98\%                  \\
                   & Overall                  & 155             & 134             & 86\%                  \\
\midrule
\multirow{3}{*}{7} & Object → Detailed Head   & 161             & 125             & 78\%                  \\
                   & Two-source Minor Head    & 76              & 73              & 96\%                  \\
                   & Overall                  & 237             & 198             & 84\%                  \\
\midrule
                   & All Volumes               & 2311            & 1952            & 84\%                  \\
\bottomrule
\end{tabular}
\caption{Validation results}
\label{tab:numvalidation}
\end{table}

Table~\ref{tab:numvalidation} presents the results for
\textbf{numerical consistency} checks for all seven volumes of the
Karnataka state finances in Year 2020--2021.  Due to the multiple
independent sources of numerical validation, errors in consistency
checks can be traced back to specific pages and rows, allowing for
targeted manual review or re-extraction.  Our pipeline prints the
locations where the specific failures occurred.

As an example, the following are the locations of failures from the Object → Detailed report on Volume~1:

\begin{verbatim}
Major_Head             2039
Description            Total 09
Page                   3
Status                 FAIL
Accounts_2018_19_Match PASS
Budget_2019_20_Match   FAIL
Revised_2019_20_Match  FAIL
Budget_2020_21_Match   PASS
\end{verbatim}

Table~\ref{tab:tedvalidation} presents results for our implementation
of \textbf{structure soundness} through the Tree Edit Distance-based
Similarity (TEDS) measure from the paper by Zhong et
al.~\cite{pubtabnet2020}.  For each of the five extracted schema
within a volume, tree structures were created for each Major Head
hierarchy.  The tree structures for two schema at the same depth were
then compared.  Since the two schema are getting generated from
different pages in the PDF document, we effectively cross-check
fiscal-head hierarchies from two different locations in the same
source. A TEDS score of zero implies that the structures are
identical. We divide the number of instances with zero TEDS against
all calculated TEDS for each volume and report that as the single
accuracy number in Table~\ref{tab:tedvalidation}.  Across all the volumes
of the Karnataka government, we have been able to achieve an accuracy
of 73\% to 96\%.

\begin{table}[h]
	\centering
	\small
	\begin{tabular}{lrr}
		\textbf{File} & \textbf{Pages} & \textbf{Accuracy} \\
		\toprule
		Volume 1      & 227            & 95.24\%           \\
		Volume 2      & 179            & 73.68\%           \\
		Volume 3      & 139            & 88.00\%           \\
		Volume 4      & 142            & 83.33\%           \\
		Volume 5      & 181            & 96.77\%           \\
		Volume 6      & 74             & 91.30\%           \\
		Volume 7      & 112            & 79.17\%           \\
		\bottomrule
	\end{tabular}
	\caption{Tree Edit Distance Similarity (TEDS)}
	\label{tab:tedvalidation}
\end{table}

We summarise the successful features of our work as three key ideas:
\begin{enumerate}
\item The ability to get accuracy metrics in the absence of ground
  truth data.
\item The ability to identify locations where the extracted data does
  not match the original PDF.  Errors at these locations can then be
  corrected with hand-labeling.  Sometimes, the extracted data can be used if failures occur in locations that are not under analysis.
\item The understanding of multi-level hierarchy during extraction,
  and corresponding validation of the hierarchy through different
  sources in the same PDF document.
\end{enumerate}

\subsection{Computation Costs}
We used Gemini 2.5 Pro for the extraction. 
Table \ref{tab:cost} presents the token count for all volumes of Karnataka’s 2020-21 finances to establish a computational baseline for replication. 
The breakdown of input, thought, and output tokens provides a granular view of the resource allocation required for this pipeline. 

\begin{table}[h]
	\centering
	\small
	 \begin{tabular}{lrrrr}
		 \textbf{File} & \textbf{Pages} & \textbf{Input Tok.} & \textbf{Thought Tok.} & \textbf{Output Tok.} \\
		 \toprule
		 Volume 1      & 227            & 1567223        & 1898941          & 443077          \\
		 Volume 2      & 179            & 1412790        & 1486710          & 430499          \\
		 Volume 3      & 139            & 977091         & 1140116          & 299388          \\
		 Volume 4      & 142            & 1052354        & 1207653          & 317168          \\
		 Volume 5      & 181            & 1452697        & 1499687          & 421164          \\
		 Volume 6      & 74             & 466169         & 583537           & 123694          \\
		 Volume 7      & 112            & 842688         & 1010741          & 235464          \\
		 \bottomrule
		 \end{tabular}
	 \caption{Gemini 2.5 Pro Token count}
	 \label{tab:cost}
	 \end{table}

\section{Future Directions}

We have demonstrated a technique for extracting information from state
fiscal documents by using a single Indian state (Karnataka) as an
example. Our research goal is to extend this work across all Indian
states and over many years of fiscal disclosures, to create a trusted
database about Indian state finance. This will require many
improvements in the implementation:

\begin{itemize}
\item \textbf{Automated meta-prompting}: Develop a pipeline where the
  LLM:
  \begin{itemize}
  \item Analyzes document structure (table types, page ranges).
  \item Generates CSV schema per table type.
  \item Creates extraction prompts automatically.
  \item Cleans up the output using a mix of rule-based and LLM-based
    cleaning.
  \item Validates and iterates until the validation passes.
  \end{itemize}
\item \textbf{Cross-state robustness}: Test across multiple Indian
  states with varying document structures.
\item \textbf{Multi-year consistency}: Ensure format stability across
  fiscal years.
\item \textbf{International applicability}: Extend to fiscal documents
  from other developing economies.
\item Create large scale reference datasets of table databases that can feed back into AI training for the future.
\end{itemize}

\subsection{Reproducible Research}

All materials, including the fiscal documents, source code, validation
scripts, extracted information and instructions are available under
MIT license at:
\url{https://github.com/xKDR/LLM_table2db}.

\section{Conclusion}

We have shown a path for LLMs to read and understand complex
hierarchical tables when provided with appropriate domain knowledge,
sequential context, and validation mechanisms. Our approach
understands the structure of 200+ page PDF documents, and extracts it
into research-ready machine-readable files with high accuracy. Our
mechanism also provides the locations of data extraction failures,
even though the ground truth is not observed.

These results suggest that LLMs do much more than regular expression
matching. They are able to leverage their vast domain knowledge
acquired in pre-training, so that prompts are able to steer
them effectively. Combined with built-in validation through summation
checks, this enables reliable automation of a task previously
requiring extensive manual effort.

While it has been applied here to Indian state finances, the
methodology is broadly applicable where complex tabular data is
published in non-machine-readable formats and where internal
consistency checks can be defined.  This opens up new avenues for
leveraging LLMs for extracting datasets from PDF files in many
situations.

\begin{acks}
We thank Yi-fan Chen and Dima Kuzmin for discussions and support. We
thank Atibhi Sharma and Meghna Saxena for their contribution to this
project.
\end{acks}

\bibliographystyle{ACM-Reference-Format}
\bibliography{paper-acm_icaif_25}

\appendix


\end{document}